\documentclass[letterpaper, 10 pt, conference]{ieeeconf}  % Comment this line out if you need a4paper

\IEEEoverridecommandlockouts                              % This command is only needed if 
\usepackage{amsmath} % assumes amsmath package installed
\usepackage{amssymb}  % assumes amsmath package installed
\usepackage{bm, balance}
\usepackage{algorithm,algpseudocode}
\usepackage[small]{caption}
\usepackage{multicol}
\usepackage{subcaption}
\usepackage{graphicx}
\usepackage{xcolor}
\usepackage[export]{adjustbox}

\usepackage[T1]{fontenc}
\usepackage[linkbordercolor={1 1 1},citebordercolor={1 1
  1},urlbordercolor={0.0 0.0
  0.0},urlcolor=blue,colorlinks=true,linkcolor=black,citecolor=black]{hyperref}
\usepackage{url}

\usepackage[backend=bibtex,style=ieee,mincitenames=1,maxcitenames=2,maxbibnames=20]{biblatex}
\title{\LARGE \bf
ROIAL: Region of Interest Active Learning \\ for Characterizing Exoskeleton Gait Preference Landscapes}

\author{Kejun Li$^{1}$, Maegan Tucker$^{1}$, Erdem B{\i}y{\i}k$^{2}$, Ellen Novoseller$^{1}$,\\ Joel W. Burdick$^{1}$, Yanan Sui$^{3}$,  Dorsa Sadigh$^{2}$,  Yisong Yue$^{1}$, and Aaron D. Ames$^{1}$ % <-this % stops a space
\thanks{This research was supported by NIH grant EB007615, NSF NRI award 1924526 and CMMI award 1923239, NSF Graduate Research Fellowship No. DGE‐1745301, and the Caltech Big Ideas and ZEITLIN Funds.}
\thanks{This work was conducted under IRB No. 16-0693.\newline  $^{1}$California Inst. of Technology, $^{2}$Stanford University, $^{3}$Tsinghua University}%
}

\begin{document}

\maketitle
\thispagestyle{empty}
\pagestyle{empty}

%%%%%%%%%%%%%%%%%%%%%%%%%%%%%%%%%%%%%%%%%%%%%%%%%%%%%%%%%%%%%%%%%%%%%%%%%%%%%%%%
\begin{abstract}

Characterizing what types of exoskeleton gaits are comfortable for users, and understanding the science of walking more generally, require recovering a user's utility landscape. Learning these landscapes is challenging, as walking trajectories are defined by numerous gait parameters, data collection from human trials is expensive, and user safety and comfort must be ensured. This work proposes the Region of Interest Active Learning (ROIAL) framework, which actively learns each user's underlying utility function over a region of interest that ensures safety and comfort. ROIAL learns from ordinal and preference feedback, which are more reliable feedback mechanisms than absolute numerical scores. The algorithm's performance is evaluated both in simulation and experimentally for three non-disabled subjects walking inside of a lower-body exoskeleton. ROIAL learns Bayesian posteriors that predict each exoskeleton user's utility landscape across four exoskeleton gait parameters. The algorithm discovers both commonalities and discrepancies across users' gait preferences and identifies the gait parameters that most influenced user feedback. 
These results demonstrate the feasibility of recovering gait utility landscapes from limited human trials.
%\ys{Concrete findings? w.r.t. science of walking}
%These results demonstrate progress towards understanding the mechanics of comfortable walking and designing gaits that more directly account for user comfort in the future.

\end{abstract}

%%%%%%%%%%%%%%%%%%%%%%%%%%%%%%%%%%%%%%%%%%%%%%%%%%%%%%%%%%%%%%%%%%%%%%%%%%%%%%%%
\section{Introduction}
Lower-body exoskeleton research aims to restore mobility to people with paralysis, a group with nearly 5.4 million people in the US alone \cite{paralysis2016}. Currently, the relationship between exoskeleton users' preferences and the exoskeleton's walking parameters is poorly understood. On the scientific front, such an understanding could yield insight into the science of walking, for instance, why exoskeleton users prefer certain gaits to others. On the direct clinical side, identifying the gaits that users prefer is critical for rehabilitation and assistive device design. Existing approaches for customizing exoskeleton walking include optimizing factors such as body parameters and targeted walking speeds \cite{wu2018individualized, ren2019personalized}, minimizing metabolic cost \cite{kim2017human, zhang2017human}, and optimizing user comfort \cite{tucker2020preference, tucker2020linecospar, thatte2018method}. More specifically, the work in \cite{tucker2020preference,tucker2020linecospar, thatte2018method} demonstrated the notion of optimizing exoskeleton gaits based on user preferences to find the optimal gait for each exoskeleton user. Learning from preferences is beneficial because it has been shown that pairwise preferences (e.g. ``Does the user prefer A or B?'') are often more reliable than numerical scores \cite{joachims2005accurately}. 

Major challenges of learning exoskeleton users' preferences include: working with limited data from time-intensive human subject experiments, ensuring user comfort and safety, accounting for user feedback reliability, and exploring the vast action space of all possible walking trajectories. Broadly speaking, methods for preference-based learning can be designed for two distinct goals.  The first is \textbf{optimization}: finding optimal gaits for specific users.  The second is \textbf{understanding}: reliably learning entire  preference landscapes.  Given the need to maximize sample efficiency from limited trials, each choice of goal implies a different sampling strategy for data collection.  %Depending on the specific goal, different action selection strategies need to be employed to maximum sample efficiency. 
Previous work \cite{tucker2020preference,tucker2020linecospar} focused on the first goal of direct function optimization, and so their approach did not reliably learn the \textit{entire} utility landscape governing user preferences across gaits. Thus, we propose an alternative approach aiming for the second goal of characterizing the entire landscape, albeit with less fine-grained data in the region close to the optimal gait. % which could not only be used to streamline the gait personalization process and facilitate the design of more comfortable gaits, but could also lead to improved exoskeleton designs that are more aligned with the mechanics of comfortable walking. 

% (the picture of online data collection for optimization of exoskeleton gaits) On the scientific front, understanding the mechanics of what makes a gait comfortable is important. On the direct clinical side, finding the gaits that the user prefers is important. Both are important. This data could be used for lots of things: directing gait generation to optimal. need to be strategic about how to collect this data - interactive data collection. Two broad paradigms: direct function optimization, the second is to map out the entire utility landscape which can be beneficial for understanding the science of walking more generally. Prior work has focused on the prior, we focus on the latter. 

 \begin{figure}[tb]
 \centering
 \includegraphics[width=\linewidth]{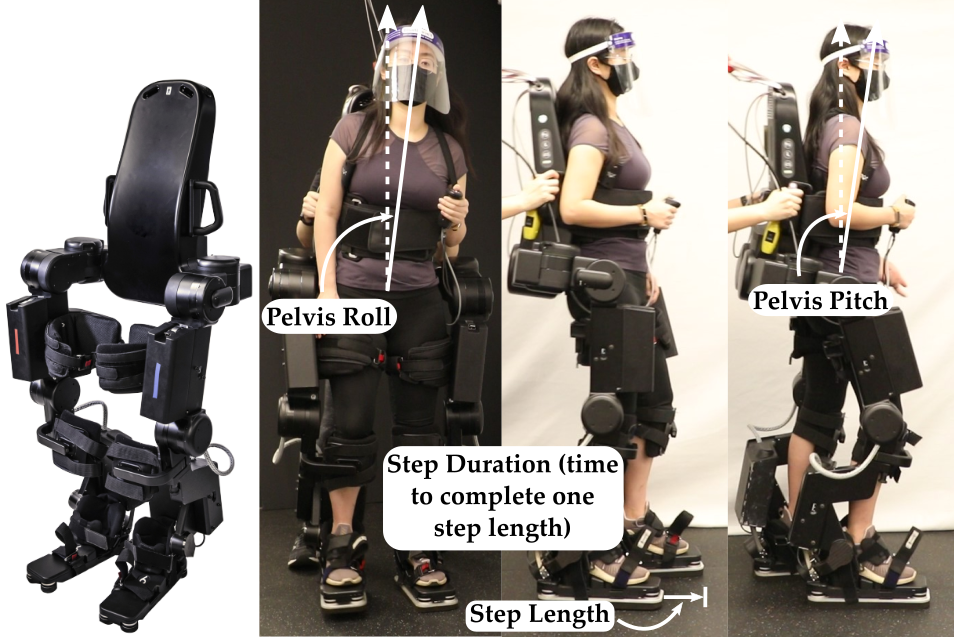}
 \caption{The Atalante exoskeleton, designed by Wandercraft, has 12 actuated joints, 6 on each leg. The experiments explore four gait parameters: step length, step duration, pelvis roll, and pelvis pitch.}
 \label{fig:Atalante}
 \vspace{-10px}
 \end{figure}

% and provide data to facilitate the design of more universally-preferred gaits in the future. 
% Previous research was aimed at locating the optimal gait parameters in as few . Previous work does not prioritize learning the entire landscape.
%Understanding the entire utility landscape would provide information towards generating more universally-preferred gaits in the future.

A consequence of exploring the entire gait parameter space is that users may be repeatedly exposed to gaits that make them feel unsafe or uncomfortable. In this work, we denote this region of undesirable gaits the ``Region of Avoidance'' (ROA) and the region of remaining gaits the ``Region of Interest'' (ROI). In prior work on the highly-related area of safe exploration  \cite{sui2015safe, schreiter2015safe,berkenkamp2016safe,sui2018stagewise}, unsafe actions are considered to be catastrophically bad and therefore must be avoided completely.  However, the resulting algorithms can be overly conservative in settings such as ours, where occasionally sampling from bad regions is tolerable.  %  but prioritizes avoidance over accurately estimating the utility function. However, in the exoskeleton walking application, it is better to only avoid bad regions once the estimate of the ROI is certain. 

This work proposes the Region of Interest Active Learning (ROIAL) algorithm, a novel active learning framework which queries the user for qualitative or preference feedback to: 1) locate the ROI, and 2) estimate the utility function as accurately and quickly as possible over the ROI. The algorithm selects samples by modeling a Bayesian posterior over the utility function using Gaussian processes and maximizing the information gain (over the ROI) with respect to this posterior. Information gain maximization for preference elicitation is a sample-efficient, state-of-the-art approach that generates preference queries that are easy for users to answer accurately \cite{houlsby2011bayesian,biyik2020active,biyik2019asking}. 
To our knowledge, our approach is the first to tackle such a region of interest active learning task.
%To our knowledge, no current framework incorporates both learning the entire utility landscape from qualitative data while also avoiding a subset of undesired actions.

The vast majority of prior work on preference learning obtains at most $1$ bit of information per preference query \cite{houlsby2011bayesian,biyik2020active, biyik2019asking,tucker2020preference, tucker2020linecospar,Xu/etal/10a,furnkranz2010preference, thatte2018method,wilde2020active,qian2015learning,sui2017multi,bengs2021preference}. ROIAL additionally learns from ordinal labels  \cite{chu2005gaussian}, which assign actions to $r$ discrete ordered categories such as ``bad,'' ``neutral,'' and ``good.'' Ordinal feedback enables ROIAL to both: 1) locate the ROI by learning the boundary between the least-preferred category (ROA) and remaining actions (ROI), and 2) estimate the utility function more efficiently within the ROI. Compared to the 1 bit of information obtained per preference, each ordinal query yields up to $\log_2(r)$ bits of information. Since ordinal feedback is identical for actions within each ordinal category, preferences provide finer-grained information about the utility function's shape within the categories.

We validate ROIAL both in simulation and experimentally.
We demonstrate in simulation that ROIAL estimates both the ROI and the utility function within the ROI with high accuracy. We experimentally demonstrate ROIAL on the lower-body exoskeleton Atalante (Fig. \ref{fig:Atalante}) to learn the utility functions of three non-disabled users over four gait parameters. The obtained landscapes highlight both agreement and disagreement in preferences among the users. Previous algorithms for exoskeleton gait optimization were incapable of drawing such conclusions; thus, this work represents progress towards establishing a better understanding of the science of walking with respect to exoskeleton gait design.

\section{Problem Statement}
We consider an active learning problem over a finite (but potentially-large) action space $\mathcal{A} \subset \mathbb{R}^d$ with $A = |\mathcal{A}|$. Each action $\bm{a} \in \mathcal{A}$ is assumed to have an underlying utility to the user, $f(\bm{a})$. The algorithm aims to learn the unknown utility function $f: \mathcal{A} \to \mathbb{R}$. The actions' utilities can be written in the vectorized form $\bm{f} := [f(\bm{a}^{(1)}),f(\bm{a}^{(2)}),...,f(\bm{a}^{(A)})]^\top$, where $\{\bm{a}^{(k)} \mid k = 1, \ldots, A\}$ are the actions in $\mathcal{A}$. Let $\bm{a}_i \in \mathcal{A}$ be the action selected in trial $i$, where $i \in \{1, \ldots, N\}$. We receive qualitative information about $f$ after each trial $i$, consisting of an ordinal label $y_i$ and (possibly) a preference between $\bm{a}_i$ and $\bm{a}_{i-1}$ for $i \ge 2$. We use $\bm{a}_{k1} \succ \bm{a}_{k2}$ to denote a preference for action $\bm{a}_{k1}$ over $\bm{a}_{k2}$, and following each trial $i$, collect these preferences into a dataset $\mathcal{D}_{p}^{(i)} = \{\bm{a}_{k1} \succ \bm{a}_{k2} \mid k = 1,2, ...,N_{p}^{(i)} \}$. Since preference feedback is not necessarily given for every trial, $N_p^{(i)} \leq i - 1$. The ordinal labels are similarly collected into $\mathcal{D}_{o}^{(i)} = \{(\bm{a}_{k},y_{k}) \mid k = 1,2,...,N_{o}^{(i)} \}$. The full user feedback dataset after iteration $i$ is defined as $\mathcal{D}_{i} := \mathcal{D}_{p}^{(i)} \cup \mathcal{D}_{o}^{(i)}$.

Ordinal feedback assigns one of $r$ ordered labels to each sampled action. These (possibly-noisy) labels are assumed to reflect ground truth ordinal categories (e.g., ``bad,'' ``neutral,'' ``good,'' etc.), which partition $\mathcal{A}$ into $r$ sets $\mathbf{O}_j$, $j \in \{1, \ldots, r\}$. We define the ROA as $\mathbf{O}_1$; for instance, in the exoskeleton setting, it consists of gaits that make the user feel unsafe or uncomfortable. Similarly, the ROA could be defined as $\bigcup_{j = 1}^n \mathbf{O}_j$ for $n > 1$, where the choice of $n$ is task-specific given the ordinal category definitions. We define the ROI as the complement of the ROA, $\mathcal{A} \setminus \mathbf{O}_1$. 

Defining $\bm{\hat{f}}_i := [\hat{f}_i(\bm{a}^{(1)}), \dots, \hat{f}_i(\bm{a}^{(A)})]^\top$ as the maximum a posteriori (MAP) estimate of the utilities $\bm{f}$ given $\mathcal{D}_i$, we aim to adaptively select the $N$ actions $\bm{a}_1, \ldots, \bm{a}_N \in \mathcal{A}$ that minimize the error in estimating $\bm{f}$ over the ROI. Defining $\bm{u} \in \{0, 1\}^A$ as a binary vector denoting which actions are within the ROI, we model the error as $\text{Error}(N) := \bm{u}^\top \lvert \bm{f}-\bm{\hat{f}}_{N} \rvert$, where the absolute value is taken element-wise. 

\section{Active Learning Algorithm}
This section describes the ROIAL algorithm (Alg. \ref{alg:ROIAL}), which leverages qualitative human feedback to estimate the ROI and utility function (code available at \cite{gitrepo}). We first discuss Bayesian modeling of the utility function, and then explain the procedure for rendering it tractable in high dimensions. We then detail the process for estimating the ROI and approximating the information gain to select the most informative actions.

\setlength{\textfloatsep}{10pt}% Decrease \textfloatsep
\begin{algorithm}[tb]
\caption{ROIAL Algorithm}
\begin{small}
\begin{algorithmic}[1]
\Require{Utility prior parameters; ordinal thresholds $b_1, \ldots, b_{r - 1}$; subset size $M$; confidence parameter $\lambda$}
\State $\mathcal{D}_0= \emptyset$, \Comment{$\mathcal{D}_i$: user feedback dataset including iteration $i$}
\State Select an action $\bm{a}_1$ at random
\State Add ordinal feedback to data to obtain $\mathcal{D}_1$
\For{i = 2,\dots, $N$}{}
 \State Update the model posterior $P(\bm{f} \mid \mathcal{D}_{i-1})$ \Comment{Eq. \eqref{eq:posterior}}
 \State Determine $\mathcal{S}^{(i)}$ by randomly selecting $M$ actions
 \State Determine $\mathcal{S}_{ROI}^{(i)} \subset \mathcal{S}^{(i)} $
\State $\bm{a}_i \gets \arg \max_{\bm{a} \in \mathcal{S}_{ROI}^{(i)}}  I(\bm{f};s_i,y_i \mid \mathcal{D}_{i-1},\bm{a})$
\State Add preference and ordinal feedback to data to obtain $\mathcal{D}_i$
\EndFor
\end{algorithmic}
\end{small}
 \label{alg:ROIAL}
 \end{algorithm}

\vspace{2mm}
\noindent \textbf{Bayesian Posterior Inference.} To simplify notation, this section omits the iteration $i$ from all quantities. Given the feedback dataset $\mathcal{D} = \mathcal{D}_{p} \cup \mathcal{D}_{o}$, the utilities $\bm{f}$ have posterior:
\begin{align} \label{eq:posterior}
    P(\bm{f} \mid \mathcal{D}_{p},\mathcal{D}_{o}) \propto P(\mathcal{D}_{p} \mid \bm{f}) P(\mathcal{D}_{o} \mid \bm{f})P(\bm{f}),
\end{align}
where $P(\bm{f})$ is a Gaussian prior over the utilities $\bm{f}$:
\begin{align*}
    P(\bm{f}) = \frac{1}{(2\pi)^{A/2} \lvert \Sigma\rvert^{1/2}} \text{exp} \left(-\frac{1}{2}\bm{f}^\top\Sigma^{-1}\bm{f}\right),
\end{align*}
in which $\Sigma \in \mathbb{R}^{A \times A}$, $\Sigma_{ij} = \mathcal{K}(\bm{a}_i,\bm{a}_j)$, and $\mathcal{K}$ is a kernel of choice. This work uses the squared exponential kernel. 

\noindent \underline{\textit{Preference feedback.}} We assume that the users' preferences are corrupted by noise as in \cite{chu2005preference}, such that:
\begin{align*}
   &  P(\bm{a}_1 \succ \bm{a}_2 \mid \bm{f}) = g_{p}\left(\frac{f(\bm{a}_1)-f(\bm{a}_2)}{c_{p}}\right), 
\end{align*}
where $g_{p}: \mathbb{R} \to (0,1)$ is a monotonically-increasing link function, and $c_{p} > 0$ quantifies noisiness in the preferences. 

% \begin{figure}[tb]
% \centering
% \includegraphics[width=0.35\textwidth]{plots/ord_noise_0.3.png}
% \caption{Ordinal likelihood for 5-category ordinal regression with ordinal noise $c_o = 0.3$. The gray dotted lines indicate divisions between categories, and each solid-color function depicts the probability of receiving a specific ordinal label given the utility $f$. }
% \label{fig:ord_prob}
% \end{figure}

\noindent \underline{\textit{Ordinal feedback.}} We define thresholds $-\infty = b_0 < b_1 < b_2 < \ldots < b_r = \infty$ to partition the action space into $r$ ordinal categories, $\mathbf{O}_1, \ldots, \mathbf{O}_r$. For any $\bm{a} \in \mathcal{A}$, if $f(\bm{a}) < b_1$, then $\bm{a} \in \mathbf{O}_1$, and $\bm{a}$ has an ordinal label of 1. Similarly, if $b_j \leq f(\bm{a}) < b_{j+1}$, then $\bm{a} \in \mathbf{O}_{j+1}$, and $\bm{a}$ corresponds to a label of $j+1$.  We assume that the users' ordinal labels are corrupted by noise as in \cite{chu2005gaussian}, such that:
\begin{align*}
    P(y \mid \bm{f}, \bm{a}) = g_{o}\left( \frac{b_{y}-f(\bm{a})}{c_{o}}\right) - g_{o}\left(\frac{b_{y-1}-f(\bm{a})}{c_{o}}\right),
\end{align*}
where $g_{o}: \mathbb{R} \to (0,1)$ is a monotonically-increasing link function, and $c_{o} > 0$ quantifies the ordinal noise.

Assuming conditional independence of queries, the likelihoods $P(\mathcal{D}_{p}\mid\bm{f})$ and $P(\mathcal{D}_{o}\mid\bm{f})$ are:
%Thus, the likelihoods $P(\mathcal{D}_{p}\mid\bm{f})$ and $P(\mathcal{D}_{o}\mid\bm{f})$ are:
\begin{align*}
    P(\mathcal{D}_{p} \mid \bm{f}) = & \prod_{k=1}^{N_p} g_{p}\left(\frac{f(\bm{a}_{k1})-f(\bm{a}_{k2})}{c_{p}}\right), \\
    P(\mathcal{D}_{o} \mid \bm{f}) = & \prod_{k=1}^{N_o} \left[g_{o}\!\left( \frac{b_{y_{k}}\!-\!f(\bm{a}_{k})}{c_{o}}\right) \!-\! g_{o}\!\left(\frac{b_{y_{k}-1}\!-\!f(\bm{a}_{k})}{c_{o}}\right)\right].
\end{align*}
Our simulations and experiments fix the hyperparameters $c_p$, $c_o$, and $\{b_j \mid j = 1, \ldots, r - 1\}$ in advance. One could also estimate them during learning using strategies such as evidence maximization, but this can be very computationally expensive, especially in high-dimensional action spaces.

Common choices of link function ($g_p$ and $g_o$) include the Gaussian cumulative distribution function \cite{chu2005preference, chu2005gaussian} and the sigmoid function, $g(x) = (1 + e^{-x})^{-1}$ \cite{tucker2020linecospar}. We model feedback via the sigmoid link function because empirical results suggest that a heavier-tailed noise distribution improves performance. We use the Laplace approximation to approximate the posterior as Gaussian: $P(\bm{f} \mid \mathcal{D}_i) \approx \mathcal{N}(\bm{\hat{f}}_i, \hat{\Sigma}_i)$ \cite{williams2006gaussian}.

\vspace{2mm} \noindent \textbf{High-Dimensional Tractability.} Calculating the model posterior is the algorithm's most computationally-expensive step, and is intractable for large action spaces. Most existing work in high-dimensional Gaussian process learning requires quantitative feedback \cite{kandasamy2015high, wang2013bayesian}. Previous work in preference-based high-dimensional Gaussian process learning \cite{tucker2020linecospar} restricts posterior inference to one-dimensional subspaces. However, the approach in \cite{tucker2020linecospar} is more amenable to the regret minimization problem because each one-dimensional subspace is biased toward regions of high posterior utility. Instead, to increase ROIAL's online computing speed over high-dimensional spaces, in each iteration $i$ we select a subset $\mathcal{S}^{(i)} \subset \mathcal{A}$ of $M$ actions uniformly at random, and evaluate the posterior only over $\mathcal{S}^{(i)}$. 

\vspace{2mm} \noindent\textbf{Estimating the Region of Interest.} Since we lack prior knowledge about the ROI, it must be estimated during the learning process. In each iteration $i$, we model the ROI as the set of actions $\{\bm{a}_k\}$ that satisfy the following criterion: $\hat{f}_{i - 1}(\bm{a}_k) + \lambda\hat{\sigma}_{i - 1}(\bm{a}_k) > b_1,$ where $\hat{\sigma}_{i - 1}(\bm{a}_k)$ is the posterior standard deviation associated with $\bm{a}_k$. The variable $\lambda$ is a user-defined hyperparameter that determines the algorithm's conservatism in estimating the ROI; positive $\lambda$'s are optimistic, while negative $\lambda$'s are more conservative in avoiding the ROA. We evaluate actions in the randomly-selected subset $\mathcal{S}^{(i)}$ and define $\mathcal{S}_{ROI}^{(i)} = \{ \bm{a} \in \mathcal{S}^{(i)} \mid \hat{f}_{i - 1} (\bm{a}) + \lambda\hat{\sigma}_{i - 1}(\bm{a}) > b_1 \}$ in each iteration 
$i$.  Note that this definition is optimistic, whereas safe exploration approaches use pessimistic definitions \cite{sui2015safe, schreiter2015safe,berkenkamp2016safe,sui2018stagewise}.

\vspace{2mm} \noindent \textbf{Action Selection via Information Gain Optimization.} To learn the utility function in as few trials as possible, we select actions to maximize the mutual information between the utility function and the preference-based and ordinal human feedback. While optimizing the entire sequence of $N$ actions is NP-hard \cite{ailon2012active}, previous work has shown that a greedy approach which only optimizes the next immediate action achieves state-of-the-art data-efficiency \cite{biyik2020active}. Hence, we adopt the same approach to solve the following optimization in each iteration $i$:
\begin{align}
    \max_{\bm{a}_{i}\in \mathcal{S}_{ROI}^{(i)}} & \:I(\bm{f} ; s_{i},y_{i} \mid \mathcal{D}_{i - 1}, \bm{a}_{i}),\label{eqn:infogain}
\end{align}
where $s_{i}$ denotes the outcome of a pairwise preference elicitation between $\bm{a}_{i}$ and $\bm{a}_{i-1}$. One can rewrite \eqref{eqn:infogain} in terms of information entropy:
\begin{align*}
    \max_{\bm{a}_{i}} H(s_{i},y_{i} \!\mid \mathcal{D}_{i - 1}, \bm{a}_{i}) \!-\! \mathbb{E}_{\bm{f}\mid\mathcal{D}_{i - 1}}\!\left[H(s_{i},y_{i} \!\mid \mathcal{D}_{i - 1}, \bm{a}_{i}, \bm{f})\right].
\end{align*}
We can interpret the first term as the uncertainty about action $\bm{a}_{i}$'s ordinal label and preference relative to $\bm{a}_{i-1}$. We aim to maximize this term, because queries with high model uncertainty could potentially yield significant information. The second term is conditioned on $\bm{f}$, and so represents the user's expected uncertainty. If the user is very uncertain about their feedback, then the action $\bm{a}_i$ gives only a small amount of information. Hence, we aim to minimize this second term. In this way, information gain optimization produces queries that are both informative and easy for users.

The second term is estimated via sampling from the Laplace-approximated Gaussian posterior $P(\bm{f}\mid\mathcal{D}_{i - 1})$. Computing the first term requires the probability $P(s_i,y_{i} \mid \mathcal{D}_{i - 1},\bm{a}_{i})$. We derive it as:
\begin{align*}
    P(s_{i},y_{i} &\mid \mathcal{D}_{i-1},\bm{a}_{i}) \\
    &= \!\int_{\mathbb{R}^A} \! P(\bm{f} \mid \mathcal{D}_{i-1},\bm{a}_{i})P(s_{i}, y_{i} \mid \mathcal{D}_{i-1},\bm{a}_{i}, \bm{f})d\bm{f}\\
    &= \mathbb{E}_{\bm{f}\mid\mathcal{D}_{i - 1}}\left[P(s_{i}, y_{i} \mid \mathcal{D}_{i - 1},\bm{a}_{i},\bm{f})\right],
\end{align*}
which we approximate with samples from $P(\bm{f} \mid \mathcal{D}_{i - 1})$.

\setlength{\textfloatsep}{20.0pt plus 2.0pt minus 4.0pt}% back to default
\begin{figure*}[tb]
     \centering
     \begin{subfigure}[b]{0.24\textwidth}
         \centering
         \includegraphics[width=\textwidth]{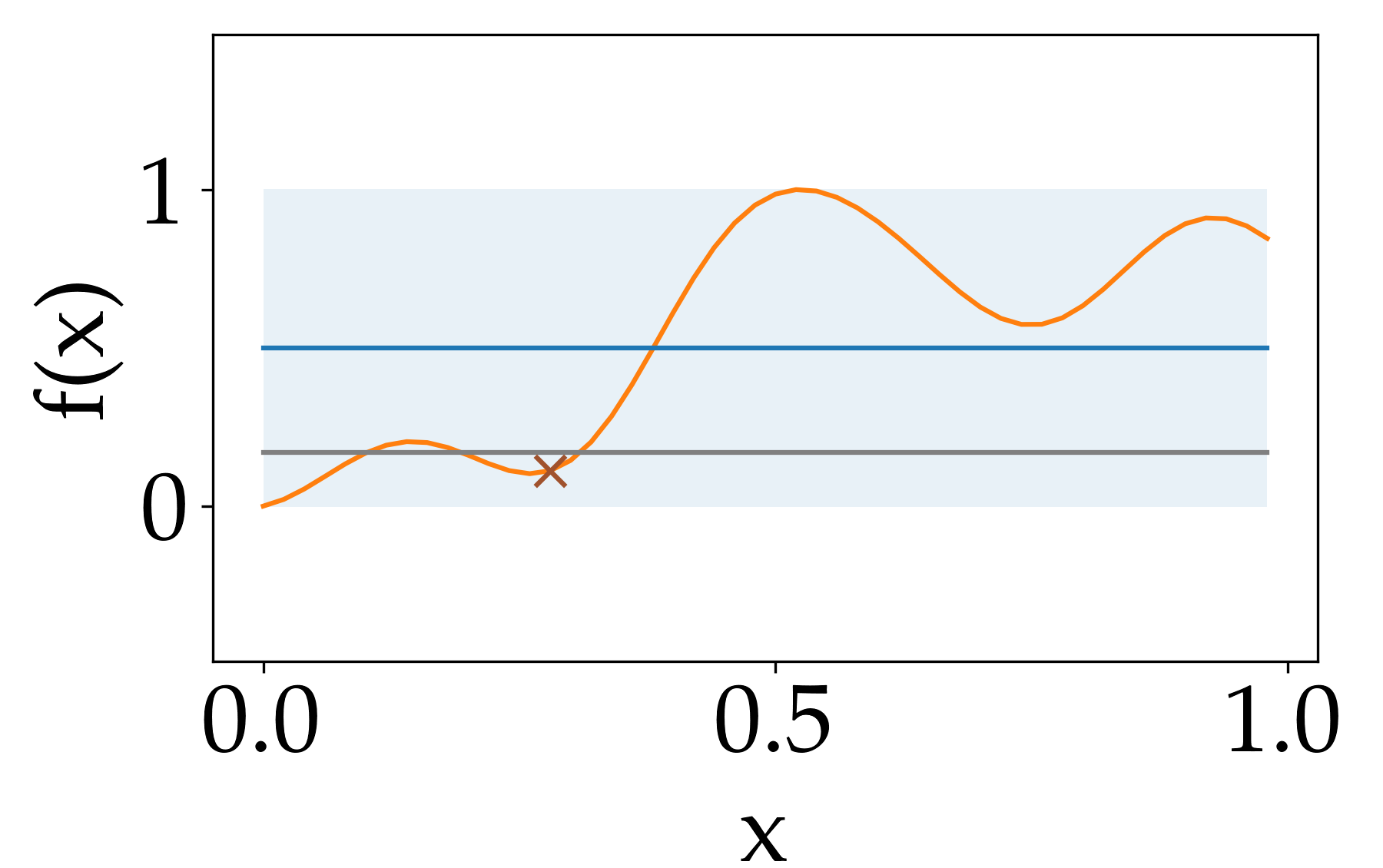}
         \caption{Iteration 1}
          \label{fig:it_1}
     \end{subfigure}
     \begin{subfigure}[b]{0.24\textwidth}
         \centering
         \includegraphics[width=\textwidth]{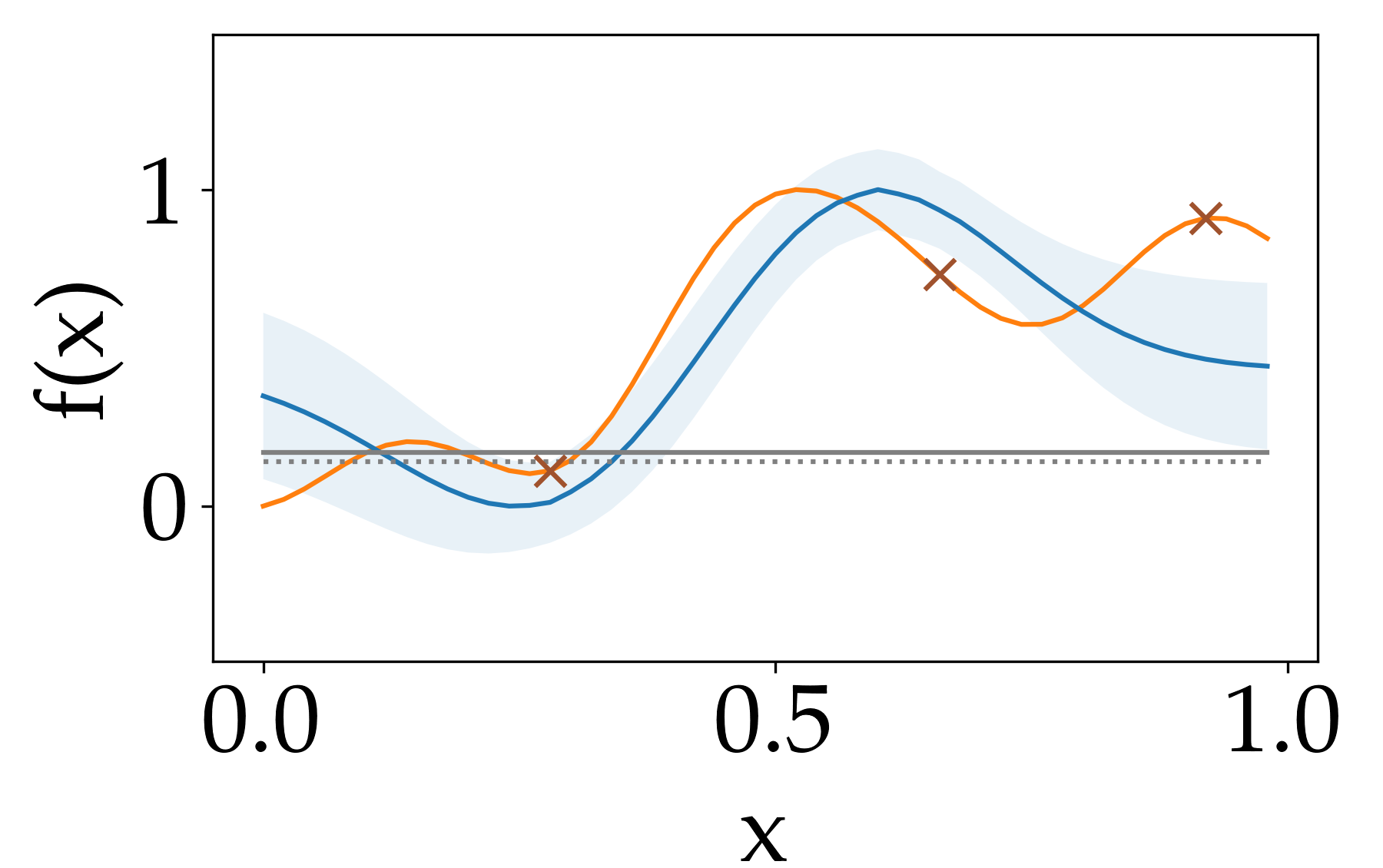}
         \caption{Iteration 3}
          \label{fig:it_3}
     \end{subfigure}
     \begin{subfigure}[b]{0.24\textwidth}
         \centering
         \includegraphics[width=\textwidth]{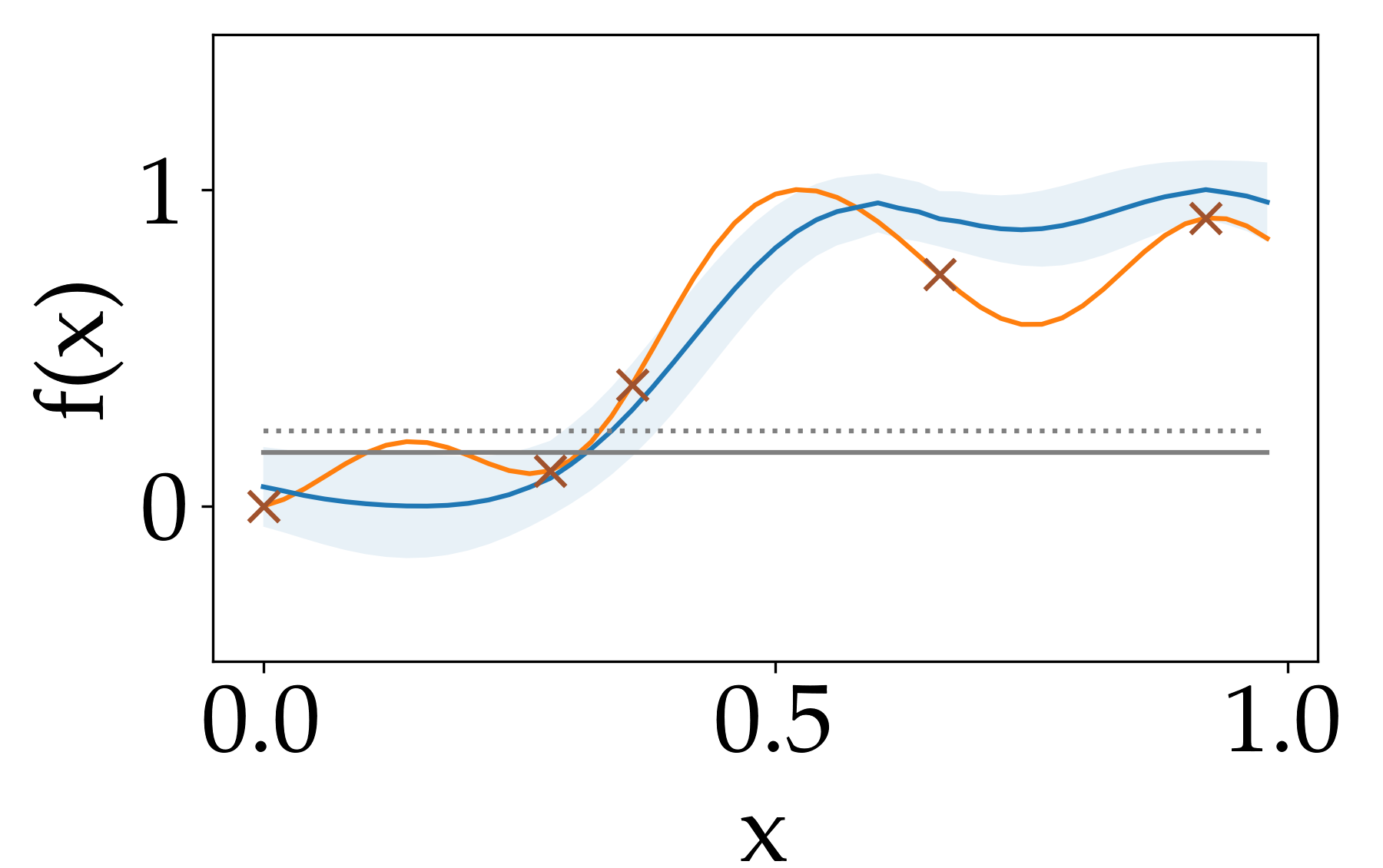}
         \caption{Iteration 5}
          \label{fig:it_5}
     \end{subfigure}
     \begin{subfigure}[b]{0.24\textwidth}
         \centering
         \includegraphics[width=\textwidth]{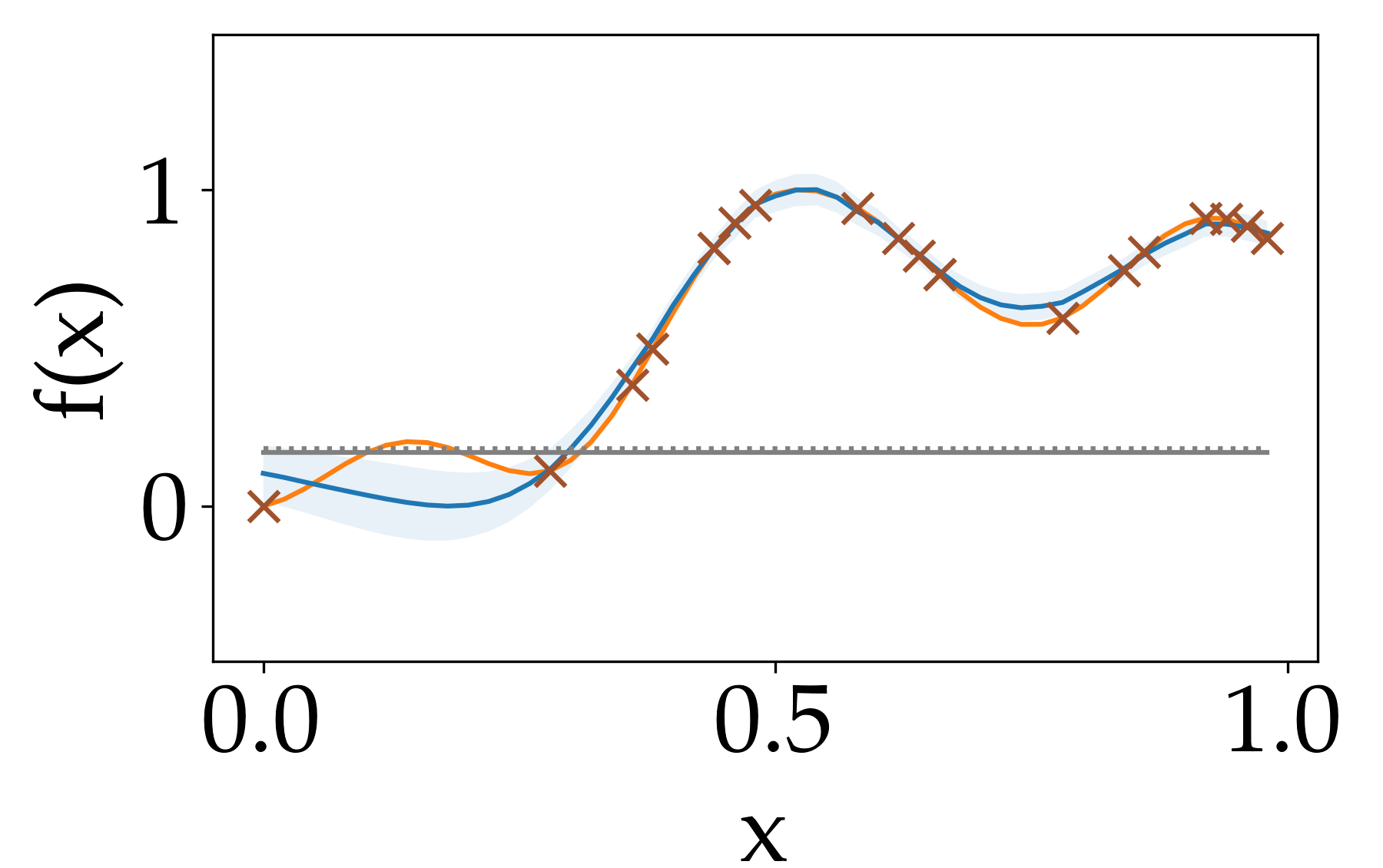}
         \caption{Iteration 20}
         \label{fig:it_20}
     \end{subfigure}
        \caption{1D posterior illustration. The true objective function is shown in orange, and the algorithm's posterior mean is blue. Blue shading indicates the confidence region for $\lambda = 0.5$. The solid grey line indicates the true ordinal threshold $b_1$: the ROI is above this threshold, while the ROA is below it. The dotted grey line is the algorithm's $b_1$ hyperparameter. The actions queried so far are indicated with ``x"s. Utilities are normalized in each plot so that the posterior mean spans the range from 0 to 1.}
        \label{fig:1dposterior}
\vspace{-3mm}
\end{figure*}

\begin{figure*}
\centering
\tabskip=0pt
\valign{#\cr
  \hbox{%
    \begin{subfigure}[b]{.21\textwidth}
    \hspace{-5mm}
    %\centering
    \includegraphics[width=\textwidth]{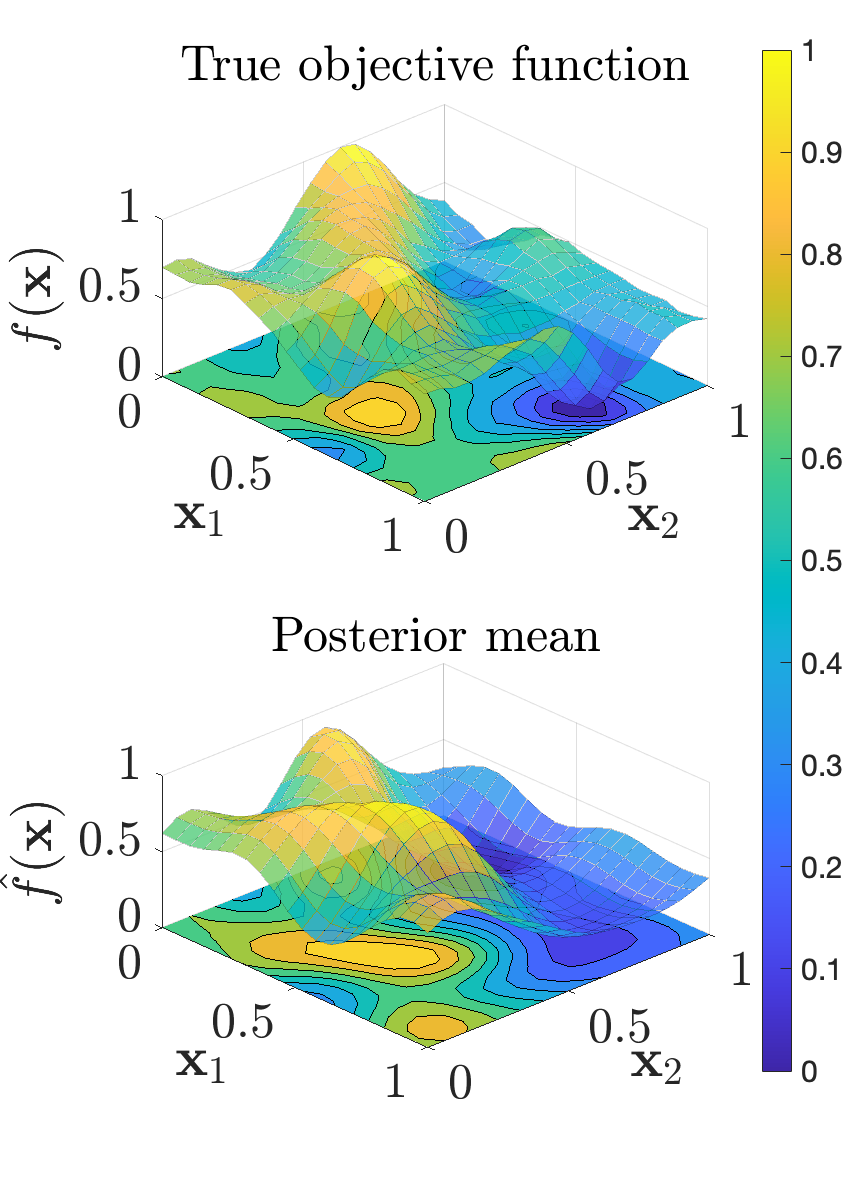}
    \caption{Synthetic function posterior}
    \label{fig:simulation_posterior}
    \end{subfigure}%
  }\cr
  \hbox{%
    \begin{subfigure}{.5\textwidth}
    \hspace{22mm}
    \includegraphics[width=\textwidth]{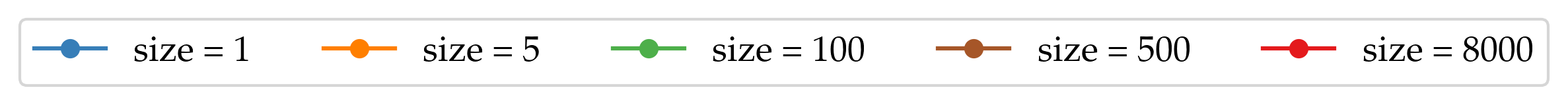}
    \end{subfigure}%
  }
  \hbox{%
    \hspace{1mm}\begin{subfigure}{.35\textwidth}
    \centering
    \includegraphics[width=\textwidth]{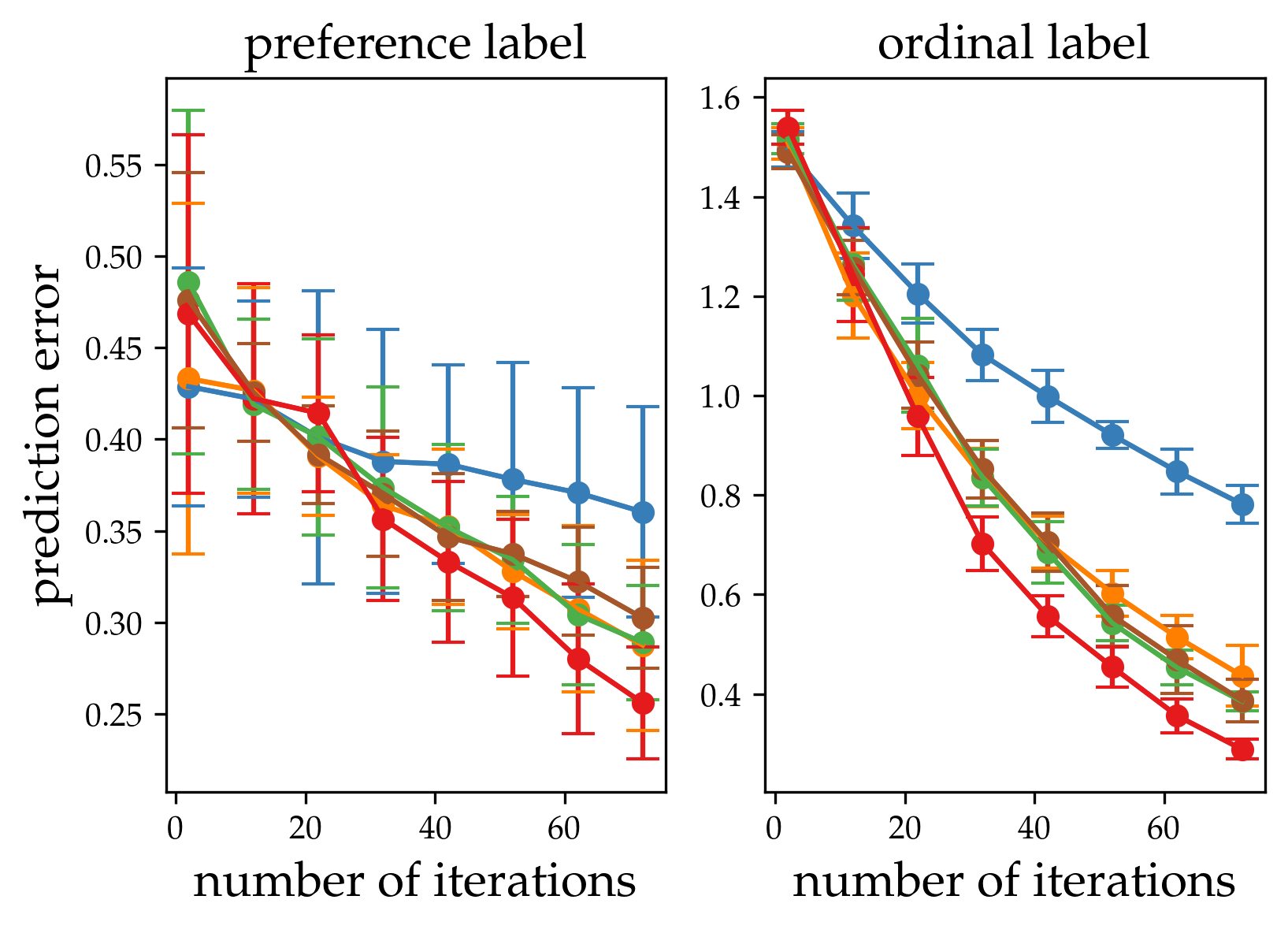}
    \caption{Hartmann3 prediction error}
    \label{fig:subset_label_hartmann3}
    \end{subfigure}%
\hspace{4mm}
  \begin{subfigure}{.35\textwidth}
    \centering
    \includegraphics[width=\textwidth]{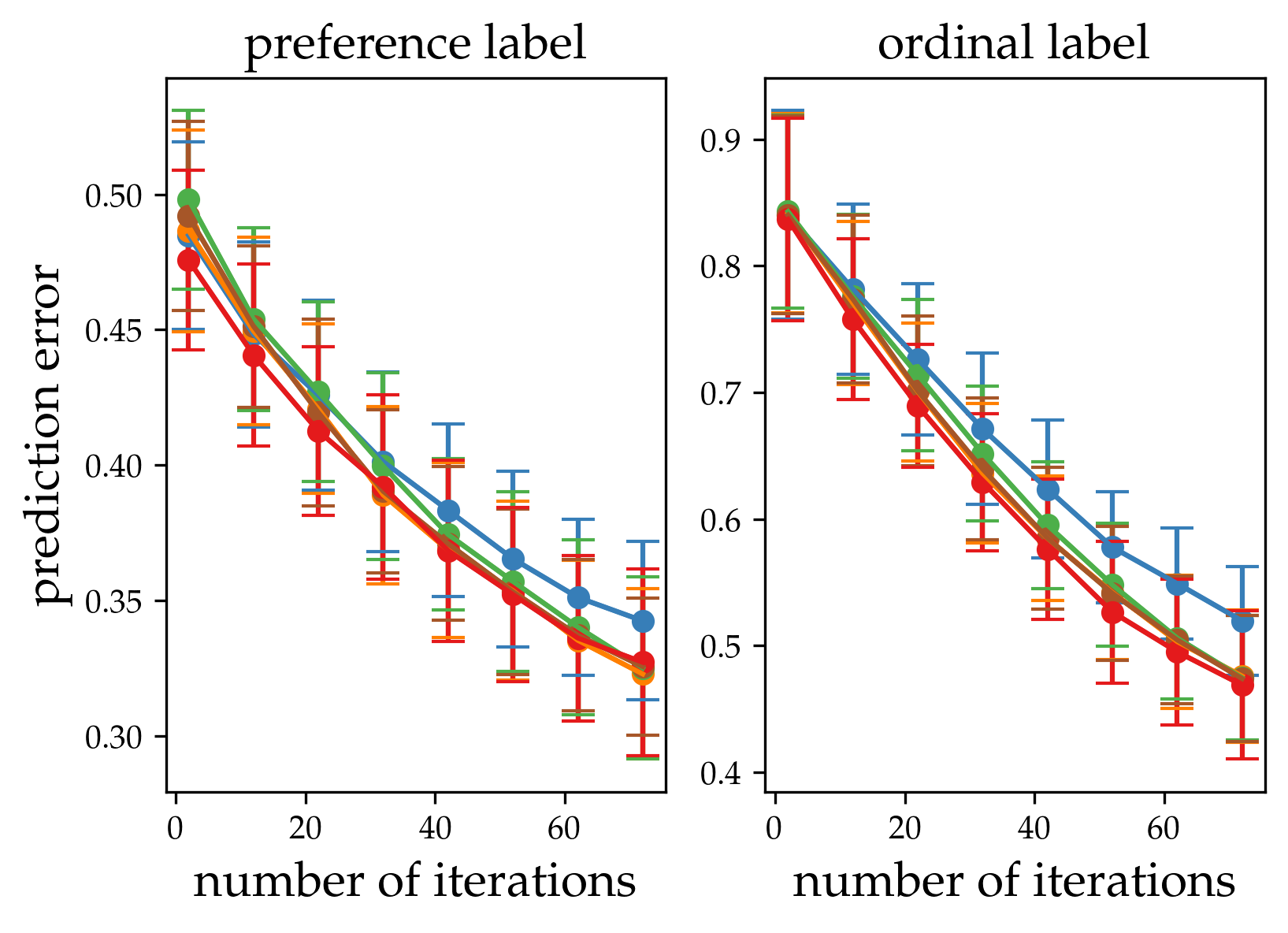}
      \caption{Synthetic function prediction error}
    \label{fig:subset_label_synthetic}
    \end{subfigure}%
  }\cr
}
\caption{Impact of random subset size on algorithm performance. a) Example 3D synthetic objective function and posterior learned by ROIAL with subset size = 500 after 80 iterations. Values are averaged over the 3rd dimension and normalized to range from 0 to 1. b-c) Algorithm's error in predicting preferences and ordinal labels (mean $\pm$ std). Each simulation evaluated performance at $1000$ randomly- selected points; the model posterior was used to predict preferences between consecutive pairs of points and ordinal labels at each point.}
  \label{fig:subset}
\vspace{-5mm}
\end{figure*}

\section{Results}
\noindent \textbf{Simulation Results.}
We evaluate ROIAL's performance on the Hartmann3 (H3) function---which is a standard benchmark for learning non-convex, smooth functions---and on 3-dimensional synthetic functions, sampled from  a  Gaussian  process  prior  over  a $20 \times 20 \times 20$ grid. As evaluation metrics, we use the algorithm's errors in preference and ordinal label prediction; these allow us to quantify performance when the true utility function is unknown. The average ordinal prediction error is defined as $\overline{\text{Error}}(N) := \frac{1}{N}\sum_{k = 1}^N |y_k^{pred} - y_k^{true}|$, and all simulations use 5 ordinal categories.\footnote{Unless otherwise stated, hyperparameters are held constant across simulations and experiments, and their values can be found in \cite{gitrepo}.}

\smallskip
\noindent \underline{\it{1D illustration of ROIAL.}} Fig. \ref{fig:1dposterior} illustrates the algorithm for a 1D objective function. Initially, ROIAL samples widely across the action space (Fig. \ref{fig:it_1}-\ref{fig:it_5}). As seen by comparing iterations 5 and 20 (Fig. \ref{fig:it_5}-\ref{fig:it_20}), the algorithm stops querying points in the ROA (actions in $\mathbf{O}_1$) because the upper confidence bound (top of the blue shaded region) there falls below the hyperparameter $b_1$ (dotted gray line).

\smallskip
\noindent \underline{\it{Extending to higher dimensions.}}
To characterize the impact of the random subset size on algorithmic performance, we compare performance of different sizes in simulation for both the H3 and synthetic functions. We calculate the posterior over the entire action space only every 10 steps to reduce computation time, and then use this posterior to evaluate the algorithm's error in predicting preference and ordinal labels. Fig \ref{fig:simulation_posterior} provides an example of a 3D posterior, Fig. \ref{fig:subset_label_hartmann3} depicts the average performance for H3 over 10 simulation repetitions, and Fig. \ref{fig:subset_label_synthetic} shows the average performance over a set of 50 unique synthetic functions. We find that a subset size of at least 5 yields performance close to using all points.

\begin{figure*}[tb]
\begin{center}
    \begin{subfigure}[b]{0.57\textwidth}
    \includegraphics[width=\textwidth]{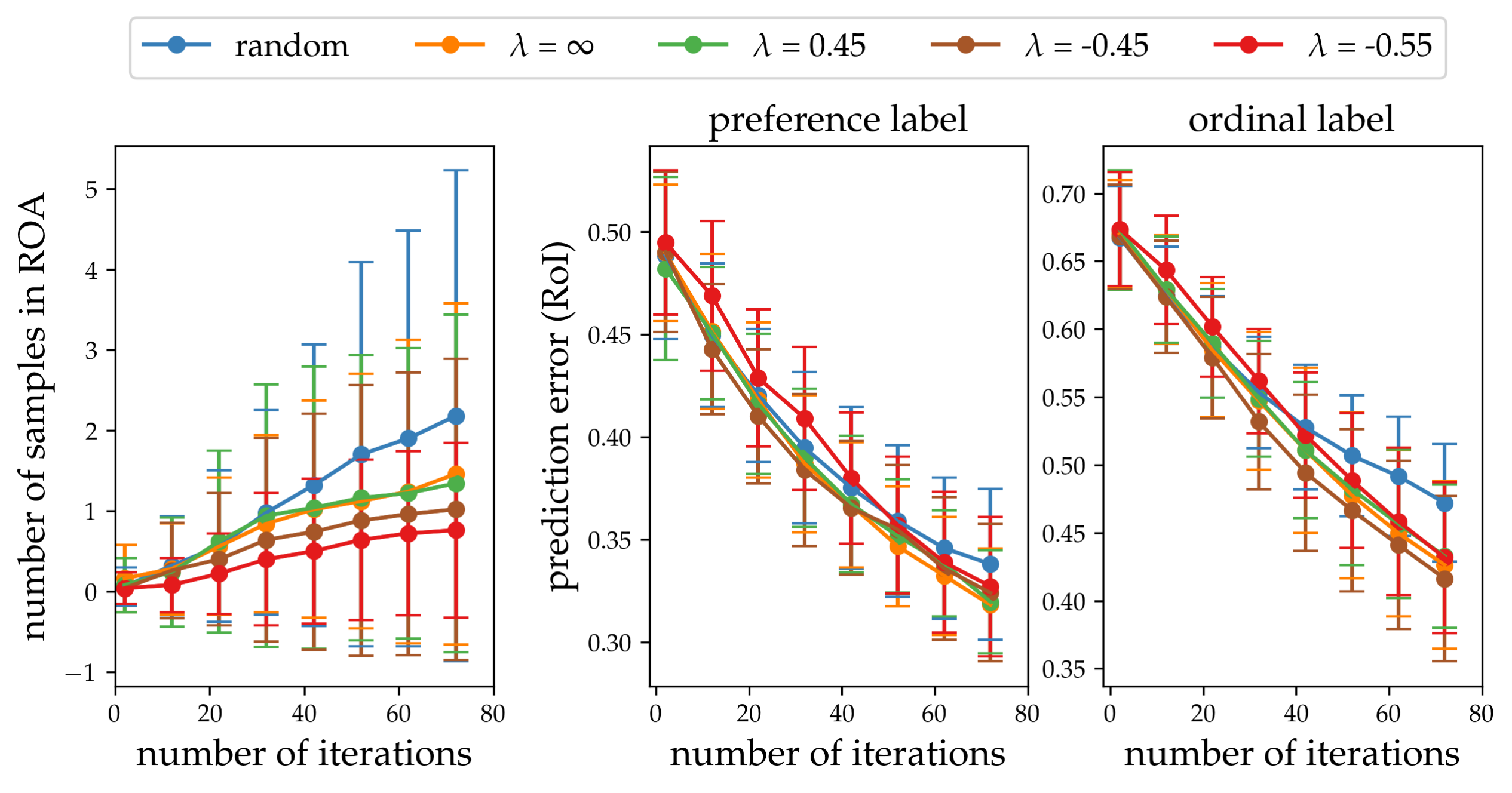}
    \caption{Number of samples in the ROA and prediction error in the ROI}
    \label{fig:ucb_idx_label}
  \end{subfigure}
  \begin{subfigure}[b]{0.4185\textwidth}
    \includegraphics[width=\textwidth]{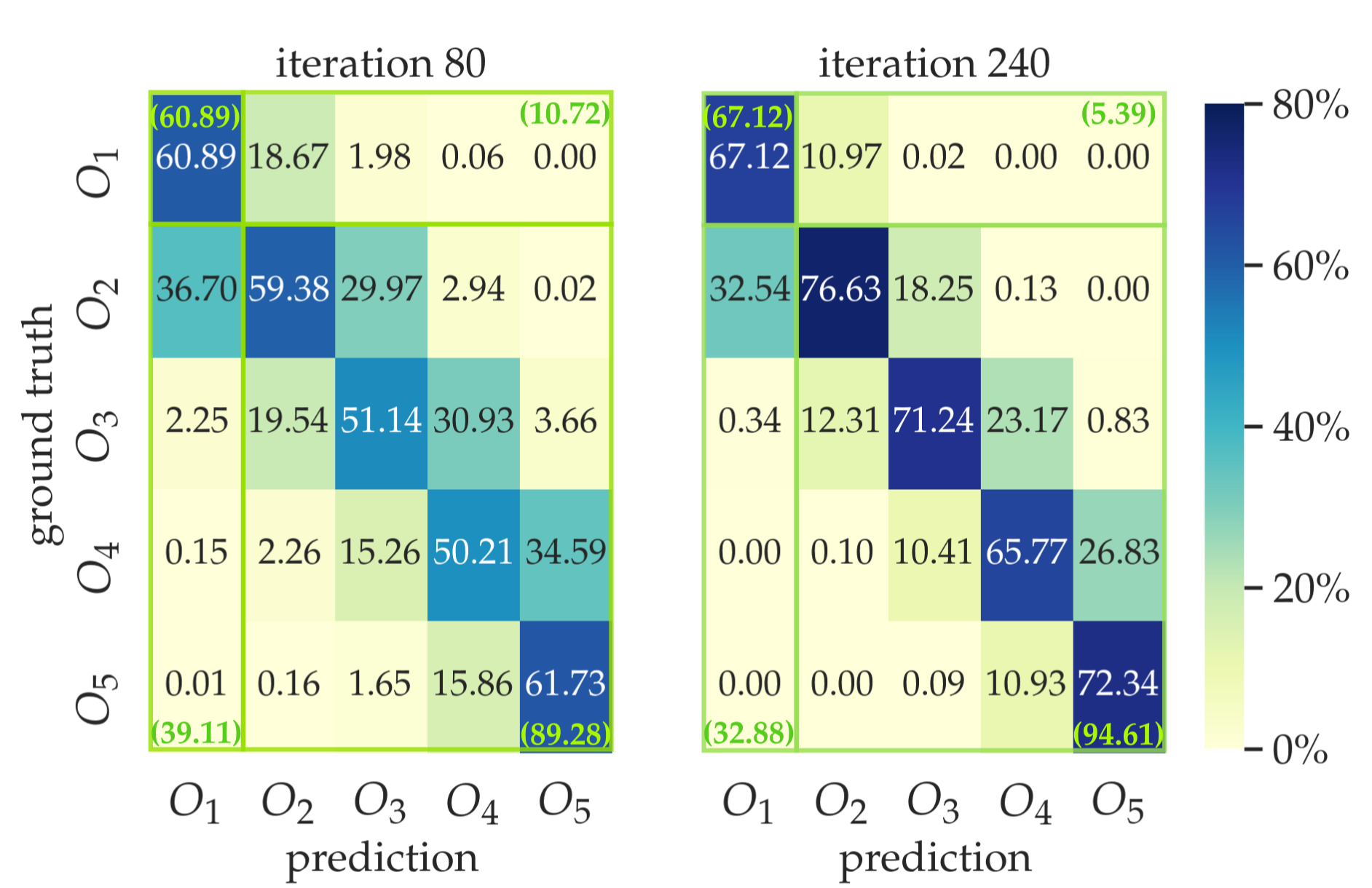}
    \caption{Confusion matrices}
    \label{fig:confusion_matrix}
  \end{subfigure}
  
  \caption{Effect of the confidence interval. All simulations are run over 50 synthetic functions with a random subset size of 500. a) Left: cumulative number of actions in the ROA ($\mathbf{O}_1$) queried at each iteration (mean $\pm$ std).  Note that as $\lambda$ increases, more samples are required for the confidence interval to fall below the ROA threshold, at which point ROIAL starts avoiding the ROA. Middle and right: error in predicting preference and ordinal labels for different values of $\lambda$; predictions are over 1,000 random actions (mean $\pm$ std). b) Confusion matrices (column-normalized) of ordinal label prediction  over the entire action space at iterations 80 and 240 with $\lambda$ = -0.45. The $2 \times 2$ confusion matrices for ROI prediction accuracy are outlined in green. Prediction accuracy increases with the number of iterations.}
  \label{fig:ucb}
\end{center}
\vspace{-8mm}
\end{figure*}

\begin{figure}[tb]
\begin{center}
    \includegraphics[width=0.455\textwidth]{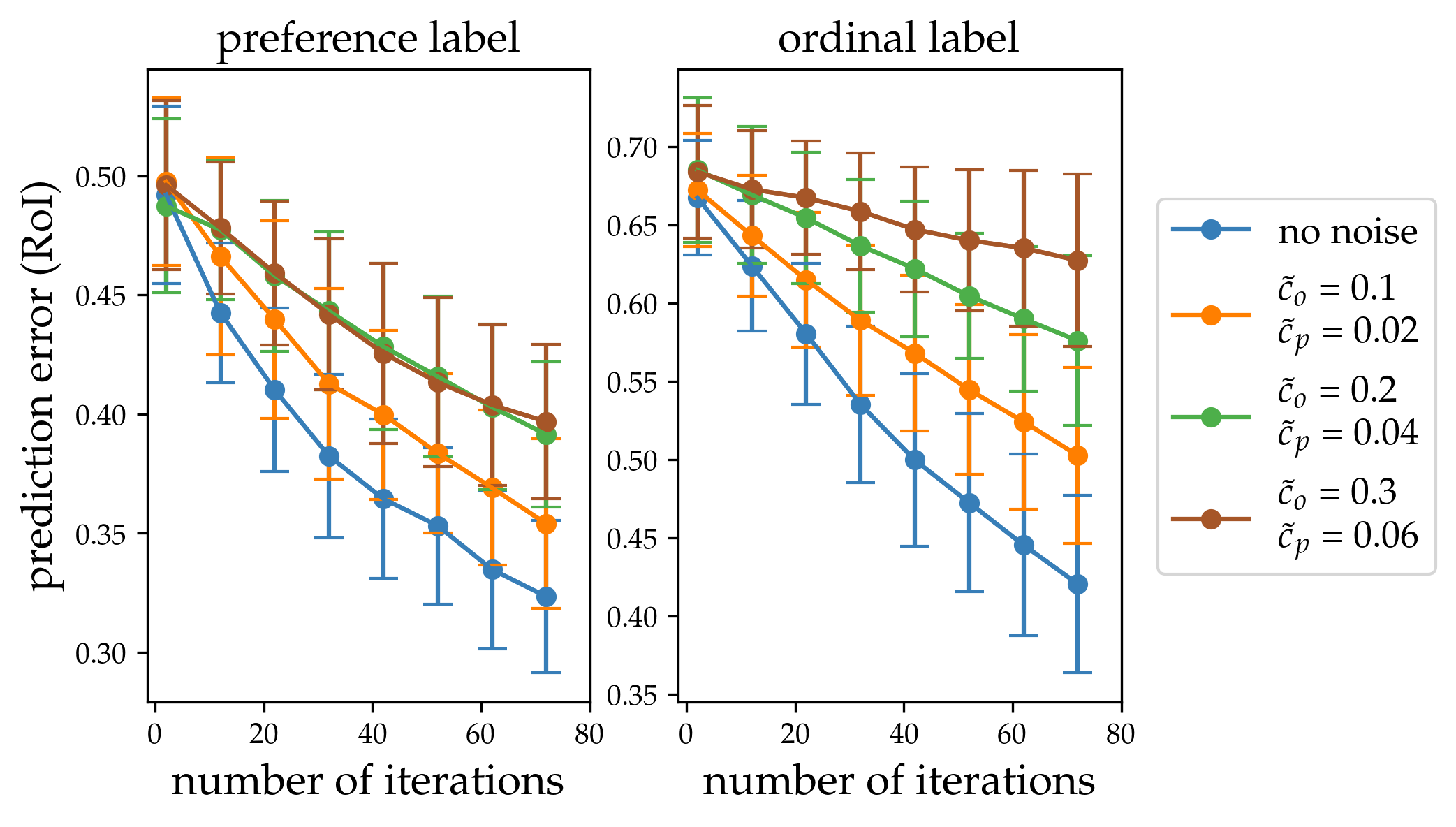}
    \vspace{-2mm}
  \caption{Effect of noisy feedback. The ordinal and preference noise parameters, $\Tilde{c}_0$ and $\Tilde{c}_p$, range from $0.1$ to $0.3$ and $0.02$ to $0.06$, respectively.  All cases use a random subset size of 500 and $\lambda = -0.45$, and each simulation uses 1,000 random actions to evaluate label prediction. Plots show means $\pm$ standard deviation.}
  \label{fig:ord_noise}
\end{center}
\vspace{-7.8mm}
\end{figure}

\vspace{1mm} 
\noindent \underline{\it{Estimating the region of interest.}}
We demonstrate the effect of the confidence parameter $\lambda$ on the number of actions sampled from the ROA and on prediction error in the ROI. Fig. \ref{fig:ucb_idx_label} demonstrates that across various values of $\lambda$, visits to the ROA decrease as $\lambda$ decreases. To confirm that restricting queries to the estimated ROI does not harm performance, we also compare label prediction error in the ROI across values of $\lambda$. When $\lambda = -0.45$, ROIAL achieves similar preference prediction accuracy and slightly-improved ordinal label prediction within the ROI compared to $\lambda = \infty$, which permits sampling over the entire action space (Fig. \ref{fig:ucb_idx_label}). Additionally, the confusion matrix (Fig. \ref{fig:confusion_matrix}) shows that the algorithm usually  predicts either the correct ordinal label  or an adjacent ordinal category. The ROI prediction accuracy (green text in Fig. \ref{fig:confusion_matrix}) indicates that ROIAL predicts whether points belong to the ROI with relatively-high accuracy. 

\vspace{1mm} 
\noindent \underline{\it{Robustness to noisy feedback.}}
Since user feedback is expected to be noisy, we evaluate the algorithm's robustness to noisy feedback generated from the distributions 
    $P(y \mid \bm{f}, \bm{a}) = g_{o}\left( \frac{\Tilde{b}_{y}-f(\bm{a})}{\Tilde{c}_{o}}\right) - g_{o}\left(\frac{\Tilde{b}_{y-1}-f(\bm{a})}{\Tilde{c}_{o}}\right)$ and  
    $P(\bm{a}_1 \succ \bm{a}_2 \mid \bm{f}) = g_{p}\left(\frac{f(\bm{a}_1)-f(\bm{a}_2)}{\Tilde{c}_{p}}\right)$ for ordinal and preference feedback, respectively, with true ordinal thresholds $\{ \Tilde{b}_j | j = 1, \dots, r-1\}$ and simulated noise parameters $\Tilde{c}_{p}$ and $\Tilde{c}_{o}$. We set $\Tilde{c}_o > \Tilde{c}_p$ because we expect ordinal labels to be noisier than preferences, as they require users to recall all past experience to give  consistent feedback, whereas a preference only involves the previous and current action. The algorithm learns more slowly with noisier feedback (Fig. \ref{fig:ord_noise}). %Fig. \ref{fig:ord_prob} depicts the likelihood for the highest simulated ordinal noise value, $c_o = 0.3$. 

\vspace{2mm} \noindent\textbf{Exoskeleton Experiments.} After demonstrating ROIAL's performance in simulation, we experimentally deployed it on the lower-body exoskeleton Atalante, developed by Wandercraft (video: \cite{video}, ROIAL hyperparameters: \cite{gitrepo}). Atalante, shown in Fig. \ref{fig:Atalante},  is an 18 degree of freedom robot designed to restore assisted mobility to patients with motor complete paraplegia through the control of 12 actuated joints: 3 joints at each hip, 1 joint at each knee, and 2 degrees of actuation in each ankle. For more details on Atalante, refer to \cite{agrawal2017first, harib2018feedback,gurriet2018towards}.

\begin{figure}[tb]
    \centering
    \includegraphics[width=0.36\textwidth]{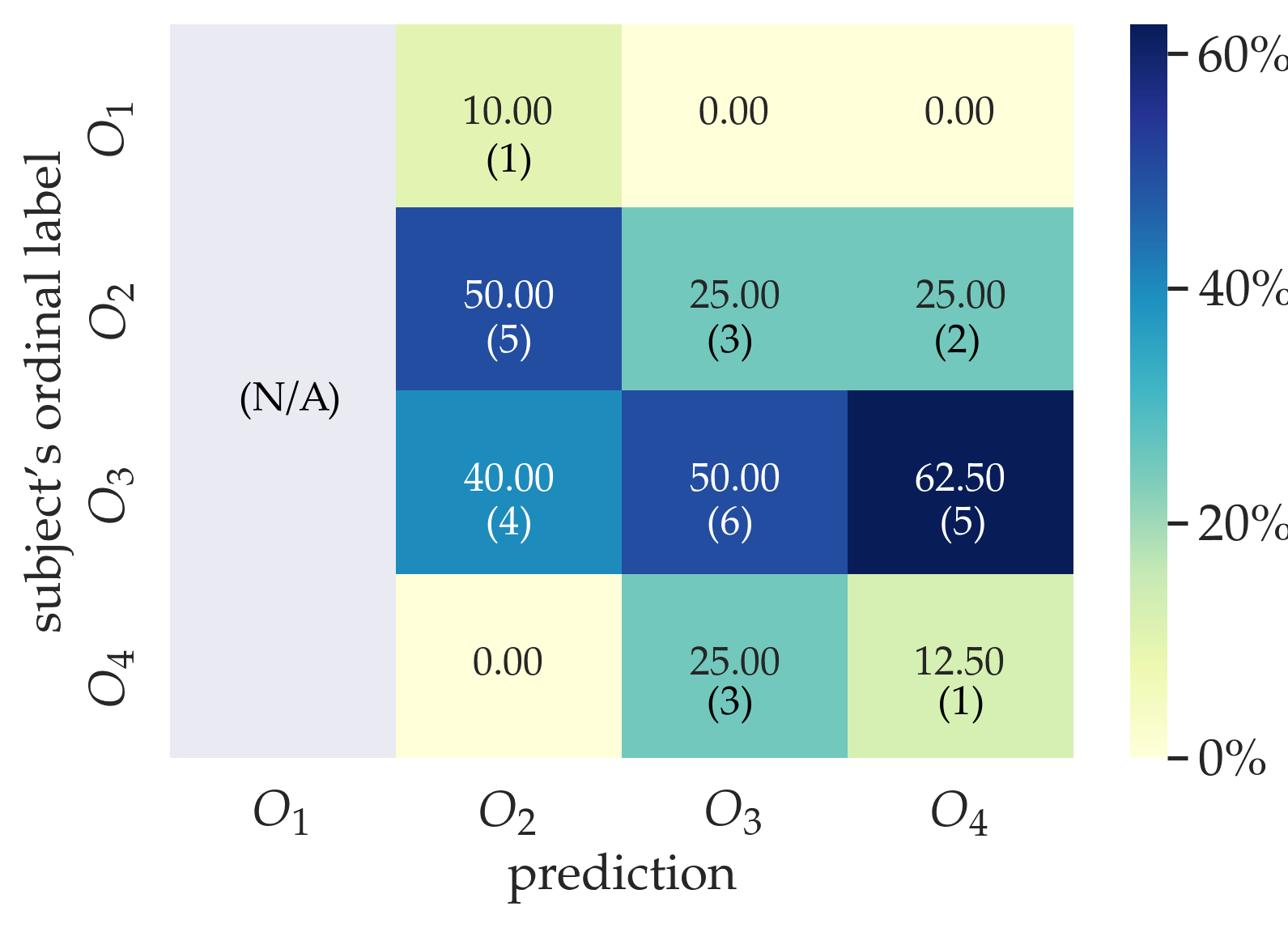}
    \vspace{-1.5mm}
      \caption{Confusion matrix of the validation phase results for all three subjects. The first column is grey because actions in the ROA ($\mathbf{O}_1$) were purposefully avoided to prevent subject discomfort. Percentages are normalized across columns. Parentheses show the numbers of gait trials in each case.}
  \label{fig:exo_confusion}
\vspace{-6.32mm}
\end{figure}

\begin{figure*}
\vspace{2mm}
    \centering
    \includegraphics[width=\textwidth]{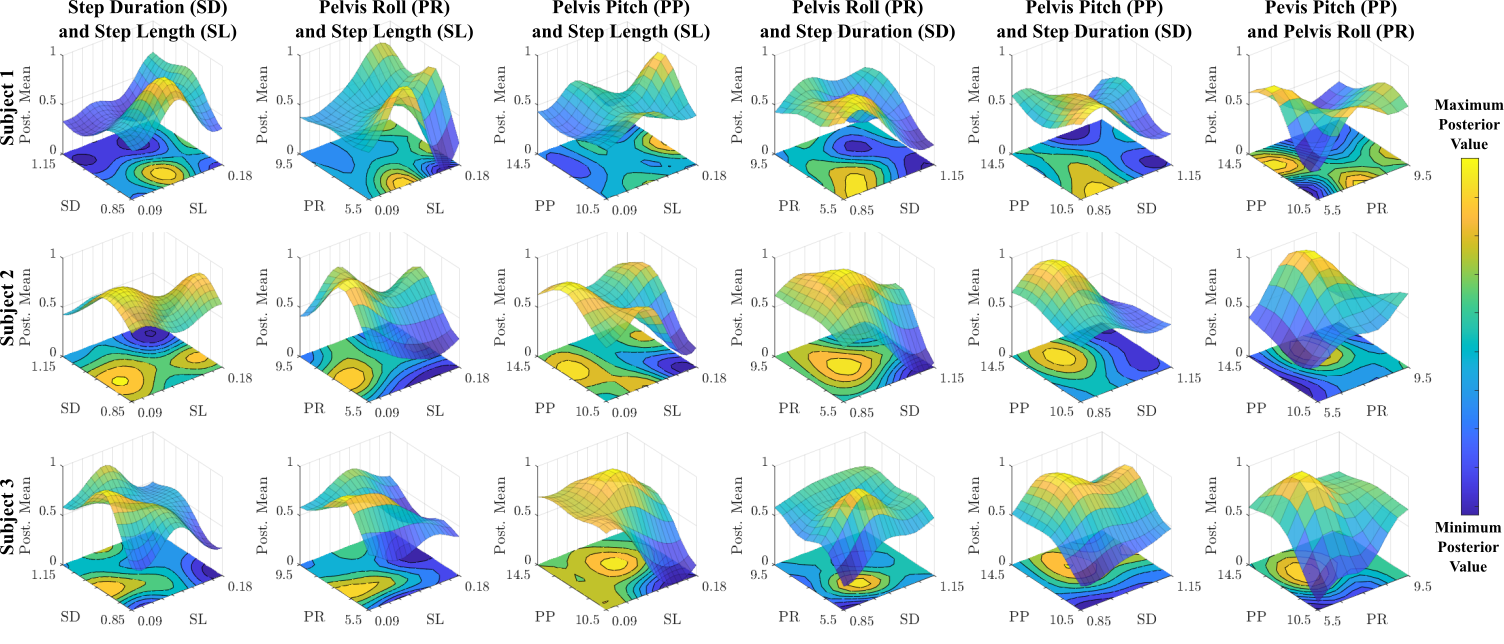}
    \caption{4D posterior mean utility across exoskeleton gaits. Utilities are plotted over each pair of gait space parameters, with the values averaged over the remaining 2 parameters in each plot. Each row corresponds to a subject: Subject 1 is the most experienced exoskeleton user, Subject 2 is the second-most experienced user, and Subject 3 never used the exoskeleton prior to the experiment.}
    \label{fig:posterior}
\vspace{-6mm}
\end{figure*}

Dynamically stable crutch-less exoskeleton walking gaits are generated through nonconvex optimization techniques (see Section II of \cite{tucker2020preference}), based on the theory of hybrid zero dynamics (HZD) introduced by \cite{ames2014human} and the HZD-based optimization method presented in \cite{hereid2018dynamic}. 
These periodic gaits are parameterized by various features, and this studies focuses on four: step length (SL) in meters, step duration (SD) in seconds, maximum pelvis roll (PR) in degrees, and maximum pelvis pitch (PP) in degrees (Fig. \ref{fig:Atalante}). These parameters were selected because exoskeleton users frequently suggested modifications to SL, SD, and PR in prior work \cite{website}, and we wanted to further study the relationship between PR and PP. We discretized these parameters into bins of sizes 10, 7, 5, and 5, respectively, resulting in 1,750 actions within a 4D action space. ROIAL randomly selected 500 actions in each iteration and used $\lambda = 0.45$ to estimate the ROI. 

The experimental procedure was conducted for three non-disabled subjects and consisted of 40 trials divided into a \textit{training phase} (30 trials) and a \textit{validation phase} (10 trials). Subjects were not informed of when the validation phase began. Subjects provided ordinal labels for all 40 gaits, and optional pairwise preferences between the current and previous gaits for all but the first trial. Four ordinal categories were considered and described to the users as:
\begin{enumerate}
    \item \textbf{Very Bad} ($\mathbf{O}_1$): User feels unsafe or uncomfortable to the point that the user never wants to repeat the gait.
    \item \textbf{Bad} ($\mathbf{O}_2$): User dislikes the gait but does not feel unsafe or uncomfortable.
    \item \textbf{Neutral} ($\mathbf{O}_3$): User neither dislikes nor likes the gait and would be willing to try the gait again.
    \item \textbf{Good} ($\mathbf{O}_4$): User likes the gait and would be willing to continue walking with it for a long period of time. 
\end{enumerate}
While including additional ordinal categories could increase the potential information gain from each query, it also increases the cognitive burden for the users and thus makes the labels less reliable. Validation actions were selected so that at least two samples were predicted to belong to $\mathbf{O}_2, \mathbf{O}_3$, and $\mathbf{O}_4$, with the remaining four validation actions sampled at random. Actions predicted to belong in $\mathbf{O}_1$ were excluded because they are likely to make the user feel uncomfortable or unsafe, and actions sampled during the training phase were explicitly excluded from the validation trials.

\vspace{1mm} 
\noindent \underline{\it{Experimental results.}} Figure \ref{fig:exo_confusion} depicts the results of the validation phase for all three subjects. These results show a reliable correlation between the predicted categories and the users' reported ordinal labels, in which the majority of the predicted ordinal labels are within one category of the true ordinal labels. 
% For instance, actions that ROIAL predicts to belong to $\mathbf{O}_2$ and $\mathbf{O}_3$ indeed most likely  belong to $\mathbf{O}_2$ and $\mathbf{O}_3$, respectively. However, ROIAL is less likely to predict actions to fall into $\mathbf{O}_1$ or $\mathbf{O}_4$ (i.e., ROIAL's predictions are regularized to the mean).  
%However, actions predicted to be in $\mathbf{O}_4$ are most likely to belong in $\mathbf{O}_3$. 
Since less than $2\%$ of the action space was explored during the experiment, we expect that the prediction accuracy would increase with additional exoskeleton trials as observed in simulation (Fig \ref{fig:confusion_matrix}). Overall, these results suggest that ROIAL can yield reliable preference landscapes within a moderate number of samples.

Figure \ref{fig:posterior} depicts the final posterior mean for each of the subjects. These utility functions highlight both regions of agreement and disagreement among the subjects. For example, all subjects strongly dislike gaits at the lower bound of PP and lower bound of PR. However, all subjects disagree in their utility landscapes across SL and SD. This type of insight could not be derived from direct gait optimization, which mostly obtains information near the optimum.

We also evaluated the effect of each gait parameter on the posterior utility using the permutation feature importance metric. The results of this test for each respective subject across the four gait parameters (SL, SD, PR, PP) are: (0.20, 0.30, 0.33, 0.27), (0.26, 0.36, 0.38, 0.29), and (0.23, 0.16, 0.21, 0.45). These values suggest that the preferences of more experienced users (Subjects 1 and 2) may be most influenced by SD and PR, while the least-experienced user's feedback may be most weighted by PP (Subject 3). The code for this test is available on GitHub \cite{gitrepo}. These results demonstrate that ROIAL is capable of obtaining preference landscapes within relatively-few exoskeleton trials while avoiding gaits that make users feel unsafe or uncomfortable.

% \vspace{1mm}
\section{CONCLUSIONS}

This work presents the ROIAL framework for actively learning utility functions within a region of interest from pairwise preferences and ordinal feedback. The ROIAL algorithm is experimentally demonstrated on the lower-body exoskeleton Atalante for three non-disabled subjects (video: \cite{video}). In simulation, ROIAL predicts utilities in the ROI while learning to stay away from the ROA. In experiments, ROIAL typically predicts subjects' ordinal labels correctly to within one ordinal category. Furthermore, the results illustrate that gait preference landscapes vary across subjects. In particular, a feature importance test suggests that the two more-experienced users prioritized step duration and pelvis roll, while a new user prioritized pelvis pitch.

Making conclusive claims about gait preference landscapes requires conducting these experiments on patients with motor complete paraplegia, as well as scaling up the experiments. Another limitation of this work is the high noise in users' ordinal labels, which may depend on factors such as prior experience and bias due to the gait execution order. Thus, future work includes designing a study to directly quantify the noise in exoskeleton users' ordinal labels. Future work also includes continuing the experiments over more trials, as prediction accuracy is expected to improve with additional data. To conclude, the ROIAL algorithm provides a principled methodology for characterizing exoskeleton users' preference landscapes in high-dimensional action spaces. This work contributes to better understanding the mechanisms behind user-preferred walking and optimizing future gait generation for user comfort.


%%%%%%%%%%%%%%%%%%%%%%%%%%%%%%%%%%%%%%%%%%%%%%%%%%%%%%%%%%%%%%%%%%%%%%%%%%%%%%%%

%%%%%%%%%%%%%%%%%%%%%%%%%%%%%%%%%%%%%%%%%%%%%%%%%%%%%%%%%%%%%%%%%%%%%%%%%%%%%%%%

%%%%%%%%%%%%%%%%%%%%%%%%%%%%%%%%%%%%%%%%%%%%%%%%%%%%%%%%%%%%%%%%%%%%%%%%%%%%%%%%
%\vspace{4mm} \noindent \textbf{Acknowledgments.}
 \section*{ACKNOWLEDGMENTS}
% \begin{small}
%\small
The authors would like to thank the experiment volunteers and the entire Wandercraft team that designed Atalante and continues to provide technical support for this project.
% \end{small}
%%%%%%%%%%%%%%%%%%%%%%%%%%%%%%%%%%%%%%%%%%%%%%%%%%%%%%%%%%%%%%%%%%%%%%%%%%%%%%%%

%\printbibliography

%\bibliographystyle{IEEEtran}
\balance
\renewcommand*{\bibfont}{\small}
\printbibliography
% \bibliography{references}

\end{document}